\documentclass{article}

\usepackage{arxiv}

\usepackage[utf8]{inputenc} 
\usepackage[T1]{fontenc}    
\usepackage{hyperref}       
\usepackage{url}            
\usepackage{booktabs}       
\usepackage{amsfonts}       
\usepackage{nicefrac}       
\usepackage{microtype}      
\usepackage{lipsum}
\usepackage{graphicx}
\usepackage{array}
\usepackage{diagbox}
\usepackage{makecell}
\usepackage{subfig}
\usepackage{multirow}
\usepackage{natbib}
\graphicspath{ {./images/} }

\title{Dynamic Domain Information Modulation Algorithm for Multi-domain Sentiment Analysis}

\author{
 Chunyi Yue \\
  School of Information and Communication Engineering\\
  Hainan University\\
  Haikou, Hainan 570228, China\\
  \texttt{d201377468@hust.edu.cn} \\
   \And
  Ang Li \\
  Department of New Networks \\
  Peng Cheng Laboratory \\
  Shenzhen, China, 518052 \\
  \texttt{lia@pcl.ac.cn}
}

\begin{document}
\maketitle
\begin{abstract}
Multi-domain sentiment classification aims to mitigate poor performance models due to the scarcity of labeled data in a single domain, by utilizing data labeled from various domains. A series of models that jointly train domain classifiers and sentiment classifiers have demonstrated their advantages, because domain classification helps generate necessary information for sentiment classification. Intuitively, the importance of sentiment classification tasks is the same in all domains for multi-domain sentiment classification; but domain classification tasks are different because the impact of domain information on sentiment classification varies across different fields; this can be controlled through adjustable weights or hyper parameters. However, as the number of domains increases, existing hyperparameter optimization algorithms may face the following challenges: (1) tremendous demand for computing resources, (2) convergence problems, and (3) high algorithm complexity. To efficiently generate the domain information required for sentiment classification in each domain, we propose a dynamic information modulation algorithm. Specifically, the model training process is divided into two stages. In the first stage, a shared hyperparameter, which would control the proportion of domain classification tasks across all fields, is determined. In the second stage, we introduce a novel domain-aware modulation algorithm to adjust the domain information contained in the input text, which is then calculated based on a gradient-based and loss-based method. In summary, experimental results on a public sentiment analysis dataset containing 16 domains prove the superiority of the proposed method.
\end{abstract}


\section{Introduction}
Sentiment classification has played a pivotal role in many scenarios, including social media monitoring, product review analysis, and customer feedback management \cite{subadi2023sentiment, jagdale2019sentiment, kumar2022aspect}. As related technologies progress, they demonstrate remarkable potential in broader applications. For instance, mental health monitoring can identify whether a person is experiencing depression or anxiety by analyzing their speech - thereby providing a reference for medical intervention \cite{andreas2023optimisation}. Additionally, for chatbots \cite{dongbo2023intelligent}, a deep understanding of user emotions can lead to more personalized and empathetic responses.

Supervised learning (SL), a widely used machine learning (ML) algorithm in natural language processing (NLP), often requires a large quantity of high-quality labeled samples. In sentiment classification, training a robust classifier for each domain also requires the labeling of a large quantity of samples, which is often difficult to obtain. Nonetheless, indiscriminately employing data from different domains to build a general classifier cannot capture the characteristics of domains. As a result, multi-domain sentiment analysis \cite{li2008multi} has been developed to alleviate the demand for domain-specific data.

Multi-domain sentiment classification benefits from multi-task learning (MTL) \cite{zhang2018overview} as it allows for the simultaneous training of classifiers on multiple domains, facilitating knowledge transfer and better, overall performance. A class of MTL models \cite{cai2019multi, ghorbanali2023exploiting, akhtar2019multi, tan2023sentiment} gives exceptional performance when doing this task. Their common feature is performing domain classification and sentiment classification tasks simultaneously, and then generating the domain information, which is transferred to the sentiment classification module. In these models, the domain classification task is considered as an auxiliary task. For sentiment classification tasks, a common expectation is to achieve the highest possible sentiment classification accuracy in each domain. However, this is not the case for domain classification tasks, where it is sufficient to provide appropriate domain information for the corresponding sentiment classification task. Clearly, the impact of domain information varies across different domains. For example, in domains that contain a lot of technical terminology, it is necessary to provide enough domain information to better understand the text. Furthermore, when the text in one domain differs significantly from the text in other domains, domain information also becomes important. This concerns a multi-task optimization (MTO) problem \cite{liu2021conflict, javaloy2021rotograd}, which focuses on the potential interference between multiple tasks.

Recent evidence shows that scalarization \cite{royer2024scalarization} is a simple and effective method for MTO, by minimizing the weighted loss of all tasks. This is also widely used in multi-domain sentiment classification models with both domain and sentiment classifiers when applying the same weight to the tasks across all domains in the simplest form. However, as mentioned before, the impact of domain classification tasks is distinct among various domains. Especially, as the number of domains increases, the search space grows exponentially, which may lead to three challenges: a tremendous requirement for computational resources, convergence problems, and high-complexity algorithms.

To respond to this issue, we propose a two-stage training strategy to dynamically balance domain classification and sentiment classification within each domain. Firstly, a multi-task neural network is selected as a basic model and a traditional grid search is used to find an appropriate hyperparameter as the weights of all domain classification tasks. Subsequently, an algorithm for adjusting domain information in an input text for sentiment classification based on learnable domain-aware modulation is introduced, while rescaling weights of the domain classification tasks is an immediate consideration. Specifically, the word embeddings of the input text are obtained first, and then a learnable modulation $\sigma$ is added to it, which can be seen as argumentation or attenuation of the domain information in the input text. It is worth noting that $\sigma$ is dynamically learned based on the input text. To figured out the $\sigma$, we propose a loss-based and gradient-based method. On the one hand, the gradient of the inputs$\nabla_x L_d$ is computed with respect to domain classification loss, and on the other hand, a learnable step size(scalar) $\lambda$ is optimized by minimizing the sentiment loss for each domain using Adaptive Moment Estimation (Adam) \cite{kingma2014adam}. They are together to determine the direction and magnitude of the $\sigma$. Furthermore, a rescaling $\lambda$ strategy based on the validation set is designed, which is crucial for ensuring the model's performance. 

The method proposed in this paper simplifies the joint hyperparameter optimization problem into finding an appropriate scalar step size for each domain while aiding in model convergence. Due to the incremental algorithm employed in our methodology, it theoretically ensures performance on any domain is not lower than that of the original model; this is because setting $\sigma$ to 0 guarantees the performance equivalent to the original model. Overall, the contributions of our work are summarized as follows:

1) We propose an approach for balancing the domain classification and sentiment classification tasks in each domain based on domain-aware modulation for multi-domain sentiment classification, simplifying the original joint optimization problem into an independent learning process of scalars in the same quantity as domains.

2) We present a hybrid gradient-based and loss-based optimization method to program the modulation, which are used to regulate the domain information contained in an input text. Their direction and magnitude are decided by the gradients of the inputs for the domain classification and step sizes learned through the achievement of sentiment classification.

3) We design a step-scale strategy based on the validation set. By contrasting the model performance with and without the modulation vectors on the validation set, the step sizes are further adjusted. If the performance deteriorates, we consider it due to an excessively large step size; if the performance remains stable, we explore a larger step size. This strategy helps inhibit overfitting.

\section{RELATED WORK}
\label{sec:relatedwork}
\paragraph{Multi-domain Sentiment Classification.}
With the advent of deep learning, neural networks have shown their advantages in multi-domain sentiment classification. Convolutional Neural Networks (CNNs) \cite{li2021survey} and Recurrent Neural Networks (RNNs) \cite{medsker2001recurrent} serve as feature extractors, albeit in different ways. Their advanced versions, including Long Short-Term Memory (LSTM) \cite{graves2012supervised}, Gated Recurrent Units (GRUs) \cite{chung2014empirical}, Residual Networks (ResNets) \cite{he2016identity}, and mixed versions further propelled the advancement in this field. For multi-domain sentiment classification, models are expected to filter out both useful shared knowledge and domain-specific knowledge. Two typical frameworks for this are: one that possesses a shared feature extraction layer and domain-specific layers, as illustrated in studies \cite{liu2016deep}, and the other that employs general frameworks with domain-related feature extraction methods - like attention mechanisms, as represented in Domain attention model (DAM) \cite{yuan2018domain}.

Transformer, introduced by Vaswani et al. \cite{vaswani2017attention}, revitalizes NLP. Unlike CNNs and RNNs, it enables parallel processing of sequence data due to self-attention, position encoding and so on, laying the foundation for the development of pre-trained language models (PLMs), such as BERT \cite{devlin2018bert} and GPT \cite{radford2018improving}. PLMs are a specific application of transfer learning in NLP, which leverages rich semantic knowledge gained from a large-scaled dataset to improve the performance of downstream tasks. In multi-domain sentiment classification, PLMs can handle domain-specific nuances while preserving strong generalization abilities across different domains, setting new benchmarks, such as\cite{roccabruna2022multi}. Different from these approaches, our proposed method (although also two-stage) does not involve any fine-tuning of model parameters. The fine-tuning and storage of a large number of model parameters not only consumes computational resources but also increases the risk of overfitting.

Recently, some ML techniques, such as active learning, adversarial learning, and contrastive Learning, are combined with neural networks to impact multi-domain sentiment analysis positively. To address data imbalance, the REFORMIST approach\cite{katsarou2023reformist} adopted hierarchical attention with Bidirectional LSTMs (BiLSTMs) framework and introduced active learning to query informative data points to achieve knowledge transfer between domains. The MUTUAL approach \cite{katsarou2023mutual} incorporates active learning to select the most informative samples through uncertainty sampling based on a stacked BiLSTM-based Autoencoder with an attention mechanism, which shows higher accuracy on public datasets with 16 different domains. In addition, DaCon \cite{dai2023dacon} employs both adversarial and contrastive learning to improve the performance of multi-domain Text Classification.

In principle, many of the aforementioned models can be taken as a base model for our research, as long as they include the sentiment classifiers and domain classifiers.
\paragraph{Hyperparameter Optimization.}
In MTL, a commonly used method to balance tasks is scalarization, which introduces constant hyperparameters as weights for each task. The choice of these hyperparameters often directly impacts the performance of MTL models. Therefore, numerous works are dedicated to hyperparameter search or learning by maximizing the performance of MTL, which is known as hyperparameter optimization \cite{feurer2019hyperparameter}.

Grid Search is a classic hyperparameter setting method, which configures hyperparameters by evaluating all possible points in a limited subset of the hyperparameter space. Yet, it is computationally expensive and impractical for high-dimensional spaces. Random Search \cite{zabinsky2009random}, on the other hand, which samples hyperparameter values from a probability distribution, is more effective at finding hyperparameters than grid search in certain situations. However, it still requires a large number of evaluations to identify the optimal set. Bayesian Optimization \cite{frazier2018tutorial} regards hyperparameter optimization as a probabilistic inference problem, using techniques such as Gaussian Processes to alternately explore and exploit the hyperparameter space. Nevertheless, as mentioned at the beginning of this work, this method is suitable for optimizing continuous domains with fewer than 20 dimensions, and needs a longer optimization time. Evolutionary algorithms, such as Genetic Algorithms (GA) \cite{holland1992genetic} and Particle Swarm Optimization (PSO) \cite{kennedy1995particle}, iteratively improve candidate solutions by simulating the process of natural selection. They are especially well-suited for complex optimization problems, but also face problems such as slow convergence. Additionally, some hybrid algorithms, such as PB2\cite{parker2020provably}, have enriched hyperparameter optimization algorithms.

Our work is also closely related to hyperparameter optimization, specifically in determining the proportion of domain classification tasks across multiple domains in sentiment classification. However, unlike the methods mentioned above, we adopt a more lightweight approach to achieve this goal, taking into account the characteristics of the task.
\paragraph{Gradient-based Approaches in ML.}
For deep neural networks, a gradient-based method, Stochastic Gradient Descent (SGD)\cite{amari1993backpropagation} is frequently employed to update the network parameters. This method aims to minimize the loss function by tuning parameters in the direction of the negative gradient. SGD updates the model parameters iteratively in a small batch of training samples with a fixed learning rate. However, this method is particularly sensitive to the learning rate, and also updates by the randomly selected data batches creates noise. Therefore, several algorithms have proposed adaptive learning rates, such as Adam, Adaptive Subgradient Methods (Adagrad) \cite{duchi2011adaptive} and, an adaptive learning rate method (Adadelta) \cite{zeiler2012adadelta}, leading to more efficient and stable convergence.

In multi-task optimization, the methods are usually divided into two categories: loss-based and gradient-based. The basic consideration for gradient-based methods is that gradients from different tasks may point in conflicting directions and cause interference. Therefore, these methods operate directly on the gradients of each task. For example, in Gradient surgery for multi-task learning (PCGrad) \cite{yu2020gradient}, the conflict between two gradients is measured by the negative cosine similarity. If two gradients are conflicting, the gradient interference can be prevented by projecting each gradient onto the normal plane of the other gradient, named ``gradient surgery''. In this paper, ``tragic triad of multi-task learning'', i.e., conflicting gradients, high curvature, and large gradient differences are demonstrated thoroughly. Gradient Sign Dropout (GradDrop) \cite{anguelov2020just} addresses the conflict issue through a random masking procedure, which is a simple and widely applicable method. Besides, noticing that naive MTL may lead to some tasks being trained more thoroughly than others,  Impartial multi-task Learning (IMTL-L) \cite{liu2021towards} proposes a more equitable way to learn multiple tasks. For task-shared model parameters, to ensure that the parameter updates have the same impact on each task, when aggregating the gradients of the original losses of all tasks ${g_t}$ into a single gradient $g$, its projections on ${g_t}$ must be equal. For task-specific parameters, all losses remain on a comparable scale. For domain-specific parameters, the scaled losses ${\alpha_t L_t}$ are enforced to be constant for all tasks. However, simply setting the scaling factors as ${\alpha_t = 1/L_t}$ causes oscillations during the learning process. Therefore, the author constructs an appropriate scaling loss function $g(s)$, allowing both the network parameters $\theta$ and the scaling parameters $s$ to be optimized by minimizing $g(s)$. These methods, based on different objectives, have designed corresponding convergent parameter rules, providing valuable references for this field.

Adversarial robustness \cite{bai2021recent} refers to the ability of machine learning models to maintain performance when encountering adversarial examples, which are intentionally designed to cause the model to make mistakes. A common approach to accomplishing robustness is through adversarial training - a min-max optimization problem. Specifically, the process involves maximizing the loss by learning perturbations of the input images to generate adversarial examples, and then minimizing the loss by training the model parameters on those examples to minimize the corresponding objective function value. The simplest way of solving inner maximization is Fast Sign Gradient Method (FSGM) \cite{goodfellow2014explaining}, which determine the update rule of perturbations of the input images through a gradient-based method, and restrict it to the surface of a norm ball. This algorithm also has some variants, such as Fast Gradient Method (FGM) \cite{miyato2016adversarial} and Projected Gradient Descent (PGD) \cite{madry2017towards}. 

Our work was inspired by the above related work and innovatively introduces modulated learning into multi-domain sentiment classification to dynamically balance the domain classification and sentiment classification tasks on each domain.
\paragraph{Federated Learning.}
Federated Learning\cite{mcmahan2017communication} was proposed to solve the practical challenge of how to effectively leverage large amounts of privacy-sensitive data collected on modern mobile devices to improve user-centric performance. Traditional deep learning models are not suitable for scenarios involving decentralized and sensitive data. To address this, Federated Learning retains data locally and trains a shared model by aggregating locally computed updates on a central server. Since the method transmits updates rather than raw data, it not only reduces communication overhead but also helps protect user privacy. Recently, a new framework called FedIT\cite{zhang2024towards} has been developed, which integrates Federated Learning with large language models (LLMs). In this framework, each client fine-tunes the model on local data using parameter-efficient tuning methods such as LoRA. These updates are then sent back to the server, which aggregates them using methods like Federated Averaging (FedAvg) to refine the shared model.

Federated Learning can also be viewed as a two-stage learning process: in the first stage, clients train models locally on their private data; in the second stage, the server aggregates the updates from multiple clients to produce a shared model. However, this class of methods differs from the approach we propose in several ways. First, due to differing objectives, in terms of structure, Federated Learning follows a two-stage process of distributed training followed by shared model aggregation, while our approach adopts the opposite order: shared model initialization followed by distributed training. Second, while Federated Learning focuses on obtaining shared model parameters, our work emphasizes adjusting the proportion of classification tasks across different domains. Additionally, while Federated Learning typically uses weighted averaging (e.g., FedAvg) to combine local results, our method introduces a novel gradient-based and loss-based mechanism to dynamically adjust the task distribution ratio across domains.
\section{BASIC MODEL}
\label{sec:basicmodel}
\subsection{Overall Framework.}
DAM is one of the most important approaches in multi-domain sentiment analysis, as it flexibly incorporates domain information into sentiment feature representations through attention mechanisms. We choose DAM as the basic model in the first stage, with slight modifications - namely, replacing the LSTM in DAM with the latest effective Extended Long Short-Term Memory (xLSTM) \cite{beck2024xlstm}, and adopting a variant method for information integration in the domain attention mechanism. This is because xLSTM is an upgraded version of LSTM, offering more powerful language processing capabilities. We call this model $BS\_mlt$, and its overall framework is shown in Figure 1, consisting of three components.

\begin{figure} 
    \centering
    \includegraphics[width=16cm]{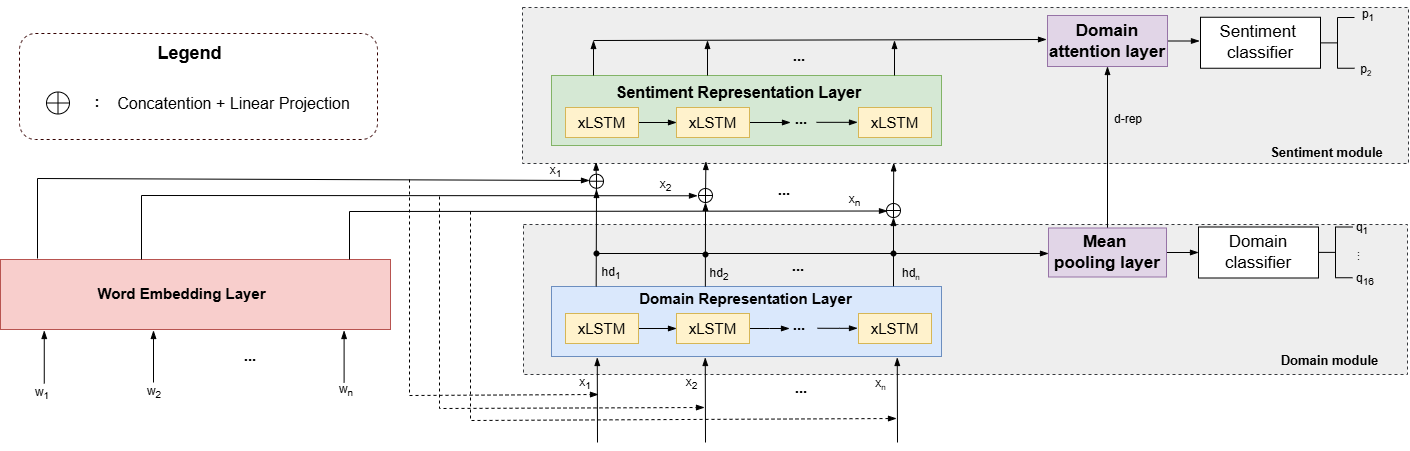}
    \caption{The basic framework of DAMA.}
    \label{fig_1}
\end{figure}

\paragraph{The word embedding layer.} The one-hot representations of the input tokens are the model's input, denoted as $W = \{w_1, \cdots, w_n\}$, where $n$ is the number of the tokens. Through an embedding layer, $W$ is transferred to the word embeddings $X = \{x_1, \cdots, x_n\}$, which are dense and semantic relationship-aware vectors.
\paragraph{The domain module.}To further capture long and short-term dependencies between tokens, the word embeddings are fed into an xLSTM to transform them into context-aware hidden states $H_d$, which then passes through a mean pooling layer, transforming into a fixed-size vector $V_d$ to represent the domain information of the input text. On the one hand, $V_d$ is the input of the domain classifier, used to predict a probability distribution over the domain labels $\{q_1,q_2,\cdots,q_{m}\}$, where $m$ is the number of domains; on the other hand, it is sent to the sentiment module to guide the classification process.
\paragraph{The sentiment module.}Similar to the domain module, the word embeddings are input into another xLSTM to generate context-aware hidden states $H_s$. The $V_d$ generated from the domain module and $H_s$, enters into the domain attention layer and is then converted to a fixed-size vector $V_s$ , a domain-aware sentiment representation of the input text. It is the input of a sentiment classifier, used to predict a probability distribution over sentiment labels $\{p_1,p_2\}$, which represent ``positive'' and ``negative'', respectively.

\subsection{Algorithms}
In this section, we introduce the two main algorithms involved in the basic model.
\paragraph{xLSTM.} this work proposes two types of cell units to replace the original ones in LSTM. In our experiments, we use the better-performing units. The variants include the introduction of exponential activation functions and a normalizer state. The implementation process is formulated as:
\begin{equation}
c_t = \mathrm{f}_t c_{t-1} + \mathrm{i}_t z_t 
\end{equation}
\vspace{-4mm}
\begin{equation}
n_t  = \mathrm{f}_t n_{t-1} + \mathrm{i}_t
\end{equation}
\vspace{-4mm}
\begin{equation}
h_t = \mathrm{o}_t \left(\frac{c_t}{n_t}\right) 
\end{equation}
\vspace{-2mm}
\begin{equation}
z_t = \varphi\left({w}_z^{\top}x_t + r_z h_{t-1} + b_z\right) 
\end{equation}
\vspace{-4mm}
\begin{equation}
\mathrm{i}_t = \exp\left({w}_{\mathrm{i}}^{\top} x_t + r_{\mathrm{i}} h_{t-1} +b_{\mathrm{i}}\right) 
\end{equation}
\vspace{-4mm}
\begin{equation}
\mathrm{f}_t = \sigma\quad\mbox{or} \quad  \exp\left({w}_{\mathrm{f}}^{\top} x_t + r_{\mathrm{f}} h_{t-1} +b_{\mathrm{f}}\right) 
\end{equation}
\vspace{-4mm}
\begin{equation}
\mathrm{o}_t = \sigma\left(w_{\mathrm{o}}^{\top} x_t + r_{\mathrm{o}} h_{t-1} + b_{\mathrm{o}}\right) 
\end{equation}
Equations (1) to (7) describe the update processes for the cell state $c_t$, an normalizer state $n_t$, hidden states $h_t$, cell input $z_t$, input gate $i_t$, forget gate $f_t$, and output gate $o_t$ in the cell. $w$ and $r$ are weights, and $\varphi$, $\sigma$ and $exp$ represent Tanh, Sigmoid and Exponential activation function, respectively. The usage of exponential activation functions in the input and forget gates is to solve the state tracking problems mentioned by Merrill\cite{merrill2024illusion} in recent work, and the normalizer state is used to normalize the cell state.

\paragraph{Domain attention.} This is an attention mechanism used to extract domain-related sentiment information from the shared feature pool. The core idea is to regard the domain representation as a query to match the features generated by the sentiment module. After attention weights on these features are generated, the weighted sum of the features are used as input to the sentiment classifier. This attention mechanism can be expressed as follows.
\begin{equation}
H_d = \mbox{xLSTM}_d(W)
\end{equation}
\vspace{-4mm}
\begin{equation}
H_{d s} = \mbox{concat}\left(H_d, W\right) 
\end{equation}
\vspace{-4mm}
\begin{equation}
V_d = \mbox{mean-pooling}\left(H_d\right) 
\end{equation}
\vspace{-4mm}
\begin{equation}
H_s = \mbox{xLSTM}_s\left(H_{d s}\right) 
\end{equation}
\vspace{-4mm}
\begin{equation}
I = \mbox{concat}\left(V_d, H_s\right) 
\end{equation}
\vspace{-4mm}
\begin{equation}
S = \mbox{ReLU}\left(W^\top I + b\right) 
\end{equation}
\vspace{-4mm}
\begin{equation}
A_{ttn} = \mbox{softmax}(S)
\end{equation}
\vspace{-4mm}
\begin{equation}
V_s = A_{ttn}^\top H_s 
\end{equation}
According to the description in equation (8) to (15),  the context-aware hidden states $H_d$ from $xLSTM_d$ in the domain module are concatenated with the word embeddings of the input $W$ element-wise, acting as the inputs $I$ of another $xLSTM_s$ to generate context-aware hidden states $H_s$ in the sentiment module. Then $V_d$, a domain vector representation of the input obtained by transforming $H_d$ through a mean-pooling layer, and $H_s$ interacts via a linear projection layer (weights $W$ and bias $b$) and ReLU activation function to produce attention scores $S$, which are transformed by Softmax activation into the probability attention distribution $A_{ttn}$ over $H_s$. Finally, the weighted sum over $H_s$ represents the sentiment vector of the input text $V_s$, used for sentiment classification. 

\subsection{Joint Training}
Joint training is adopted to optimize the parameters of the base model in our work by designing a joint loss function. Specifically, the joint loss $L_\theta$ is calculated as:
\begin{equation}
L_\theta = \sum_{j=1}^m L^s_\theta\left(p\left(\hat{y}_s^j \mid x^j\right), y_s^j\right) +\gamma\sum_{j=1}^m  L^d_\theta\left(p\left(\hat{y}_d^j \mid x^j\right) \cdot y_d^j\right)
\end{equation}

The first term on the right side of the equation is the sentiment classification error $L^s$, and the second term is the domain classification error $L^d$. The coefficient $\gamma$ is used to determine the weight of the domain classification task. Because for multi-domain sentiment classification tasks, domain classification is an auxiliary task, $\gamma$ is typically set between 0.0 and 0.1. Besides, $m$ is the number of domains, $y_s^j$, $\hat{y}_s^j$, $y_d^j$ and $\hat{y}_d^j$ represent the true and predicted sentiment labels, as well as the true and predicted domain labels in the $j^{th}$ domain, respectively; and $\theta$ refers to all parameters to be optimized in the neural network. 

\section{METHODOLOGY} 
\label{sec:methodology}
\subsection{Domain-aware Modulation Algorithm (DAMA)}
To dynamically fit the impact of domain classification tasks for sentiment classification tasks in all domains, a direct approach is to assign a weight to each domain. In this case, the loss function is defined as Equation (17):
\begin{equation}
L_\theta= \sum_{j=1}^M L_\theta\left(p\left(\hat{y}_s^j \mid x^j\right), y_s^j\right) \nonumber +\sum_{j=1}^M {\gamma_j L_\theta\left(p\left(\hat{y}_d^j \mid x^j\right), y_d^j\right)} 
\end{equation}
\vspace{-5mm}
\begin{equation}
\sum_{j=1}^M {\gamma_j L_\theta\left(p\left(\hat{y}_d^j \mid x^j\right), y_d^j\right)} = {\gamma\left(L_{\delta^j}\left(p\left(\hat{y}_d^j \mid x^j+\delta^j\right), y_d^j\right)\right.}
\end{equation}
\vspace{-5mm}
\begin{equation}
~~~~~~~~~~~~~~~~~~~~~~~~~~~~~~~~~~~~~~~~~~= {\gamma\left(L _ { \theta } \left(p\left(\hat{y}_d^j \mid x^j\right)\right)\right)+\Delta^j}
\end{equation}

Due to the difficulty in efficiently finding the optimal combination of $\gamma_j$ values, we reformulate this problem by adding a modulation term $\sigma^j$ in the input as equation (18) to modify the original domain classification loss as equation (19). $\sigma^j$ is sample-dependent, that is, for the $i^{th}$ input, the corresponding modulation is $\sigma^{ij}$. For simplicity,  the subscript $i$ is omitted. $\Delta^j$ in equation (19) denotes the change in the loss function introduced by incorporating $\sigma^j$. 

$\sigma$ is a vector that can be added to the input, and its addition should cause a change in the loss. Therefore, one of the straightforward methods is computing the gradient $g:=\nabla_{\delta_j}L_d$ of the input with respect to domain classification. $\sigma$ varies along this direction, domain classification loss will increase or decrease. Then, $g$ is multiplied by a learnable step size $\lambda^j$. Because $\lambda^j$ is the only learned parameter, loss-based method by minimizing the domain-specific sentiment classification loss is feasible. The classical Adam algorithm can be used to update it. Collectively, $\sigma$ is written as:
\begin{equation}
\delta_j=\lambda^j \nabla_{\delta_j} L_{\delta j}\left(p\left(\hat{y}_d^j \mid x_j\right), y_d^j\right)
\end{equation}
Once $\lambda^j$ is learned for each domain, we can inference whether the domain classification loss should be decreased or increased by its sign and the gradient.
\subsection{\texorpdfstring{$\lambda$ scaling} strategy}
Although only a scalar need to be learned for each domain, there is still a risk of overfitting. Hence, a $\lambda$ scaling strategy based on the validation set is designed, as summarized in Algorithm 1. 
\vspace{-5mm}
\begin{center} 
\rule{\textwidth}{0.4pt} 
\textbf{Algorithm 1}~~~~~~$\lambda$ Scaling strategy
\rule{\textwidth}{0.4pt} 
\vspace{-5mm}
\end{center}
\vspace{-5mm} 
\begin{enumerate}
  \item Require: $\lambda, \alpha_x, \alpha_{x+}$
  \item Require: $S = [\lambda], A_{x+} = [\alpha_{x+}], L_{x+} = [l_{x+}]$
  \item Require: $count = 0$, $max\_{value} = 500$
  \item Repeat:
    \item ~~~~~if {$\lambda \leq  max\_{value}$} then
    \item ~~~~~~~~~~ $count += 1$
    \item ~~~~~~~~~~~if {$A_{x+}[-1] < \alpha_x$} then
        \item ~~~~~~~~~~~~~~~~~if {$\alpha_x$ in $A_{x+}$} then
            \item ~~~~~~~~~~~~~~~~~Find the first occurrence position of $\alpha_x$, $i$
            \item ~~~~~~~~~~~~~~~~~ $\lambda = S[i]$
            \item ~~~~~~~~~~~~~~~~~break
            \item ~~~~~~~~~~~else
            \item ~~~~~~~~~~~~~~~~~$\lambda = S[-1]/\alpha$
    \item ~~~~~else if {$A_{x+}[-1] = \alpha_x$} then
        \item ~~~~~~~~~~ if the elements in $A_{x+}[-t:]$ equal then
           \item ~~~~~~~~~~~~~~~ if elements in $L_{x+}[-t:]$ not equal then
                \item ~~~~~~~~~~~~~~~~~~~~ Find the pos. of min. in $L_{x+}[-t:]$, $j$
                \item ~~~~~~~~~~~~~~~~~~~~ $\lambda = S[j]$
                \item ~~~~~~~~~~~~~~~~~~~~ 
                break
        \item ~~~~~~~~~~~~~~~ else
        \item ~~~~~~~~~~~~~~~~~~~~ $\lambda = \lambda \cdot \beta$
        \item ~~~~~~~~~~ else
        \item ~~~~~~~~~~~~~~~~~break
    \item ~~~~~ Add $\lambda$ to $S$
    \item ~~~~~ Compute $\alpha_{x+}, l_{x+}$ based on $\lambda$
    \item ~~~~~ Add $\alpha_{x+}$ to $A_{x+}$
    \item ~~~~~ Add $l_{x+}$ to $L_{x+}$
\item Until $count = N_T$ 
\end{enumerate}
\vspace{-5mm}
\rule{\linewidth}{0.4pt}

In algorithm 1, $\lambda$ is the step size, and $a_x$, $a_{x+}$ represent the sentiment classification accuracy when the inputs are $x$ and $x+\sigma$, respectively. The lists S, Ax, Ax+ are used to store $\lambda$, $a_x$ and $a_{x+}$. $count$ records the number of $\lambda$ adjustments. When it reaches $N_T$, the scaling process ends. This strategy includes three conditional checks: 

If the learned $\lambda$ improves the sentiment classification accuracy $a_{x+}$ on the validation set, it is considered an appropriate value and does not require adjustment.

If $a_{x+}$ on the validation set remains the same, $\lambda$ maybe too small, causing an insufficient impact on the classification accuracy. Therefore, it is multiplied by a scaling factor $\beta>1$. At the same time, we still specify a maximum number of scaling iterations $t$. If after $t$ times scaling, the sentiment classification accuracy is not change, we then check the sentiment classification loss $l_{x+}$ on the validation set and select the $\lambda$ corresponding to the minimum loss as the final $\lambda$.

When the sentiment classification accuracy on the validation set decreases after introducing modulation, we speculate the $\lambda$ value is too large, interfering with the original sentiment classification. At this point, the first thing that needs to be determined is whether there have been any previous moments when $a_{x+}$ was equal to $a_x$. If so, we select the first occurrence as the index and use the corresponding $\lambda$, otherwise, divide $\lambda$ by a scaling factor $\alpha>1$ to reduce the $\lambda$ value. 

It should be noted that $\alpha$ and $\beta$ are set to different values to prevent the model from jumping at the same location.

\section{EXPERIMENTS}
\label{sec:experiments}
To evaluate the effectiveness of the proposed method in improving sentiment classification performance, we conduct sufficient experiments on a public dataset with 16 domains (mtl-dataset). 
\subsection{Experimental Setup}
\paragraph{Datasets.} the mtl-dataset consists of the Amazon sentiment analysis dataset, Movie Review dataset(MR), and  Internet Movie Database (IMDB), released by Liu \cite{liu2017adversarial}. Each domain contains a training set (approximately 1600 samples), a test set (400 samples), and a large amount of unlabeled data. We use the labeled data and extract 20\% samples from the training set as the validation set for each domain. The data in different domains are reviews about a specific type of products.

\paragraph{Hardware \& Software configurations.} the experiments were conducted on a computer equipped with an Intel Core i7 processor, 64GB of RAM, and an NVIDIA GeForce RTX 1080 GPU, running the Ubuntu operating system. The software environment mainly includes the programming language Python v3.9.19, and deep learning (acceleration) libraries such as PyTorch v1.8.0, CUDA v12.1, and cudnn v8.9.2. These configurations enable our experiments to proceed smoothly.

\paragraph{Basic setup.} the training process in our  method consists of two phases. In the first phase, we used GloVe to generate 300-dimensional word embeddings and set the length of input sentences as 512. The neural network's hidden states and output vectors are 300-dimensional and 64-dimensional, respectively. Considering domain classification in sentiment analysis is an auxiliary task, the shared task weight for all domains was searched by a grid search in the range [0.0-0.1] with a step of 0.02. We also applied dropout in classification layers with a drop probability of 0.5 to prevent overfitting. When optimizing parameters, we employed the Adam algorithm with a learning rate of 1e-3 and a weight decay of 1e-4. Besides, we selected the batch size and the number of epochs as 64 and 5. 

In the second phase, the parameters that need to be trained are around a dozen $\lambda$. To avoid the impact of the modulation introduced by $\lambda$ exceeding the original domain classification task weights, it is constrained within a symmetric range $[-b,b]$. By observing the gradient ratio across classification tasks, we set $b$ within the range [100-500] and use a grid search with a step size of 100 to find an appropriate $\lambda$ for each domain. In the optimization process, the Adam algorithm with a learning rate of 1e-3 was adopted and the number of epochs was set to 1. When scaling $\lambda$, $\alpha$ and $\beta$ in algorithm 1 are assigned values of 1.5 and 2, to implement bolder exploration and more cautious contraction. Other settings, such as batch size, remain consistent with the first phase.

All models involving model parameter training designate the epoch to 5, meaning 5 training epochs on all training data. In the second phase of DAMA training, the epoch is set to 1. This is because the only parameter to learn in DAMA is the step size $\lambda$, and simplifying the training objective helps the learning process converge more quickly. During the experiments, it was also observed that the training loss significantly decreased in the first epoch, and additional training did not result in substantial changes.

\subsection{Baseline and Proposed Models}
We compared the performance of our proposed model with several baseline models, which include traditional machine learning models as well as state-of-the-art deep learning techniques. Below, a brief introduction of these models is provided.

\paragraph{BS\_{all}:} A model using samples in all domains indiscriminately to train a general sentiment classifier with a simple BiLSTM architecture.

\paragraph{BS\_{single}:} A model using samples in each domain to train a domain-specific classifier, which has the same architecture with BS\_{all}. 

\paragraph{BS\_{mtl}:} A model introduced in section 3.1, which is a classic multi-task learning model with a domain module and a sentiment module that be jointly trained. This model is also referred to as the general model in our work, which serves as the basic model and accomplish multi-domain sentiment classification tasks based on a general framework.

\paragraph{BS\_{ftg}:} A model has the same architecture as BS\_mtl, but after jointly training the model using data from all domains, the sentiment classifiers are fine-tuned using data from each domain.

\paragraph{MUTUAL\cite{katsarou2023mutual}:} A multi-domain sentiment classification method via Uncertainty Sampling.

\paragraph{DaCon\cite{dai2023dacon}:} A multi-domain text classification model using domain adversarial contrastive learning.

\paragraph{DAMA:} Our proposed model, which has the same architecture and basic setup with BS\_{mtl}. 

\paragraph{BS\_scale:} A naive loss-based multi-task learning method, where $\gamma^j$ in Equation (17) is jointly trained with the other parameters in the neural network.

\paragraph{BS\_PCGrad1:} A gradient-based multi-task learning method using PCGrad, treating domain and sentiment classification tasks in different domains as different tasks (16 tasks).

\paragraph{BS\_PCGrad2:} A gradient-based multi-task learning method using PCGrad, treating domain classification tasks and sentiment classification tasks as different tasks (2 tasks).

Among them, BS\_all, BS\_single, BS\_mtl, BS\_ftg, BS\_scale, BS\_PCGrad1, BS\_PCGrad2 and DAMA follow the same experimental settings, while the experimental results for MUTUAL and DaCon are taken from the corresponding literature. The $\ast$ under some models indicates the parameters used are those optimal parameters on the test set, rather than the validation set in the first stage. 

\subsection{Experimental Results}
The overall sentiment classification accuracies of all the models are shown in Table \ref{tab:table1}.
\begin{table*}
\caption{Experimental results on baselines and DAMA \label{tab:table1}}
    \centering
\renewcommand\arraystretch{1.2}
\begin{tabular}{|c|p{1.25cm}<{\centering}|p{1.2cm}<{\centering}|p{1.2cm}<{\centering}|p{1.2cm}<{\centering}|p{1.2cm}<{\centering}|p{1.2cm}<{\centering}|p{1.2cm}<{\centering}|p{1.2cm}<{\centering}|p{1.2cm}<{\centering}|} \hline
\diagbox[width=2cm,height=1.2cm]{Datasets}{Model} & BS\_single & BS\_all & BS\_mtl & \makecell[c]{BS\_mtl\\$\ast$} & MUTUAL & DaCon & BS\_ftg & DAMA & \makecell[c]{DAMA\\$\ast$} \\ \hline
Books & 70.0 & 86.8 & 89.0 & 88.0 & 86.3-2.2 & 87.8+3.2 & 89.0-1.5 & \textbf{89.0+0.3} & \textbf{88.0+0.3} \\ \hline
DVD &	79.8 &	84.0 &	85.8 & 88.0 & 85.0+1.3 & 89.5-0.7 & 85.8+1.0 & \textbf{85.8+0.5} & 88.0+0.0 \\ \hline
Electronics & 80.3 & 87.0 & 87.3 & 87.0 & 85.3+0.5 & 91.6-2.8 & 87.3-0.8 & 87.3+0.0 & \textbf{87.0-0.3}\\ \hline
Apparel & 80.0 & 84.5 & 87.5 & 88.3 & 84.2+1.1 & 87.0+3.3 & 87.5+0.0 & \textbf{87.5+0.3} & 88.3+0.0 \\ \hline
Software & 81.8 & 88.3 & 90.3 & 90.5 &	89.5-4.0 & 90.9+0.4 & 90.3+0.0 & 90.3+0.0 & \textbf{90.5+0.3} \\ \hline
Camera & 83.0 & 88.5 & 87.3 & 87.5 & 82.7+6.3 & 93.5-1.2 & 87.3+1.8 & \textbf{87.3+0.3} & \textbf{87.5+0.3} \\ \hline
Health & 84.8 & 89.0 & 89.5 & 91.5 & 86.0+1.6 & 90.4+1.9 & 89.5+0.0 & \textbf{89.5+0.3} & \textbf{91.5+0.3} \\ \hline
Magazine & 84.0 & 88.5 & 90.3 & 90.8 &	86.8+1.8 & 94.5-2.5 & 90.3+3.0 & 90.3+0.0 & \textbf{90.8+0.3} \\ \hline
MR & 71.8 & 68.5 & 71.8 & 73.8 & 72.0+8.0 & 77.1-1.1 & 71.8+2.3 & \textbf{71.8+0.3} & \textbf{73.8+0.7} \\ \hline
Baby & 83.5 & 88.8 & 88.8 & 88.8 &	87.5-1.2 & 92.0+0.5 & 88.8+0.8 & 88.8+0.0 & 88.8+0.0 \\ \hline
Kitchen & 84.3 & 87.5 & 90.3 & 88.3 &	84.5-1.0 & 90.8-0.3 & 90.3-1.5 & \textbf{90.3+0.3} & \textbf{88.3+0.5} \\ \hline
Music & 82.3 & 83.3 & 85.3 & 84.0 & 80.0+1.9 & 86.9+1.9 & 85.3-1.8 & \textbf{85.3+0.5} & \textbf{84.0+0.3} \\ \hline
Sports & 80.3 & 87.0 & 88.3 & 89.0 & 83.3+3.7 & 91.2+0 1 & 88.3-0.5 & \textbf{88.3+0.5} & 89.0+0.0\\ \hline
Toys & 86.3 & 89.3 & 88.3 & 89.3 &	86.4-2.6 & 90.0+1.0 & 88.3-1.0 & \textbf{88.3+0.5} & \textbf{89.3+0.3} \\ \hline
IMDb & 83.5 & 84.8 & 86.3 & 87.0 &	86.8-2.1 & 88.5+3.3 & 86.3-1.3 & 86.3+0.0  & \textbf{87.0-0.8} \\ \hline
Video & 84.5 & 84.8 & 86.3 & 87.8 & 86.5-0.7 & 88.8+1.7 & 86.3-0.5 & \textbf{86.3+0.3} & \textbf{87.8+1.0} \\ \hline
Average & 81.3 & 85.7 & 87.0 & 87.5 &	84.6+0.7 & 89.4+0.5 & 87.0+0.0 & 87.0+0.3 & 87.5+0.2 \\ \hline
\end{tabular}
\end{table*}

\textbf{BS\_single, BS\_all vs. BS\_mtl:} From Table \ref{tab:table1}, it is clear that BS\_single has the poor overall performance, with an average classification accuracy of only 81.3\%. This is because each domain classifier can only utilize a small amount of samples from the specific domain. Next is BS\_all, which, although using all the data to acquire shared knowledge, performs poorly in some domains due to the lack of domain-specific features; for example, in MR, the classification accuracy is only 68.5\%, significantly lower than that of other models. The BS\_mtl model, based on multi-task learning, outperforms the two previous models due to its ability to extract both shared features and domain-specific features, achieving a classification accuracy of 87.0\%.

\paragraph{MUTUAL, DaCon vs. DAMA:} 
MUTUAL and DaCon are two methods that incorporate cutting-edge deep learning techniques into the multi-domain sentiment classification task. Their classification accuracies are shown as the original model's test accuracy plus the fluctuation brought by the introduction of new techniques. Overall, both of these improvement methods demonstrate better performance compared to initial versions. Nevertheless, for individual domains, they cannot guarantee non-degrading performance. For instance, in MUTUAL, while the classification accuracy in the MR domain improved by 8.0\%, the accuracy in the Software domain decreased by 4.0\%. Yet, the improvement in classification performance in one domain often cannot offset the decline in another, as each domain classification task is crucial.

\paragraph{BS\_mtl, MUTUAL and DaCon vs. DAMA:} 
DAMA, our proposed method, illustrates improved performance compared to the original model BS\_mtl in 11 domains (displayed in bold), with the same performance in 5 domains. The average classification accuracy across 16 domains improved by 0.3\%. We analyzed that the main reason for the stagnation in the Electronics, Software, Magazine, Kitchen, and IMDb domains is that the shared task weight of 0.02 is a relatively suitable value within the defined search range. In contrast to MUTUAL and DaCon, a major advantage of DAMA is its ability to minimize the risk of performance degradation in any domain. More importantly, our method can build upon many existing improvements. 

\paragraph{BS\_mtl (test\_set) vs. DAMA (test\_set):} 
To further validate the performance of DAMA, we examined whether DAMA could still improve the model's performance after BS\_mtl reached its performance ceiling (i.e., the highest classification accuracy on the test set) under various hyperparameter combinations. The BS\_mtl (test\_set) and DAMA (test\_set) columns show the comparative experimental results of the two models on the test set. Although there is a performance decline in the Electronics and IMDb domains, the improvements were observed in a greater number of domains. Overall, DAMA(test\_set) still outperforms BS\_mtl(test\_set). We speculate the reason for the performance degradation in some domains is mainly due to two reasons: first, BS\_mtl (test\_set) presents the optimal results from a variety of hyperparameter combinations, while DAMA (test\_set) shows the test results under only one hyperparameter setting. Second, it is related to the stopping conditions of the iterative scaling strategy in the second stage. In theory, as long as the step size is sufficiently small, it is possible to obtain at least equal test accuracy to BS\_mtl (test\_set).

\paragraph{BS\_ftg, BS\_mtl vs. DAMA:} 
In DAMA, the training process consists of two stages. At the first stage, a general model BS\_mtl is trained, while at the second stage, the sentiment classifier for each domain is improved individually. A method that also employs a two-stage training process, BS\_ftg, fine-tunes the model parameters using samples from each domain based on the general model. According to the results presented in Table \ref{tab:table1}, although the fine-tuning approach gives significant performance improvements in the Magazine and MR domains, it degrades performance in most other domains. And, the average test accuracy across all domains remains at 87.0\%, which is comparable to the performance of BS\_mtl. Compared to DAMA, BS\_ftg updates a large number of parameters to better fit specific domain data, as opposed to updating a scalar parameter, is more prone to overfitting. Additionally, since these parameters, unlike $\lambda$ in DAMA, do not have a clear interpretation—such as a step size—it is difficult to understand the combined effect of these parameter updates. That lead to a challenge to ensure non-degradation of performance and model interpretability.

\subsection{Hyperparameter Selection}
In the two-stage training process of DAMA, several important hyperparameters need to be preset, as they play a crucial role in ensuring high sentiment classification accuracy.

\begin{figure*}
    \flushleft
    \subfloat[On the validation set]{
        \includegraphics[width=8.6cm]{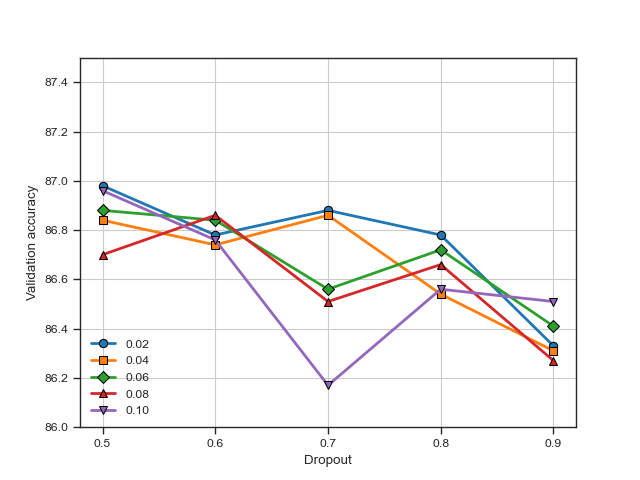}
    }
    \subfloat[On the test set \label{fig:sub2}]{%
        \includegraphics[width=8.6cm]{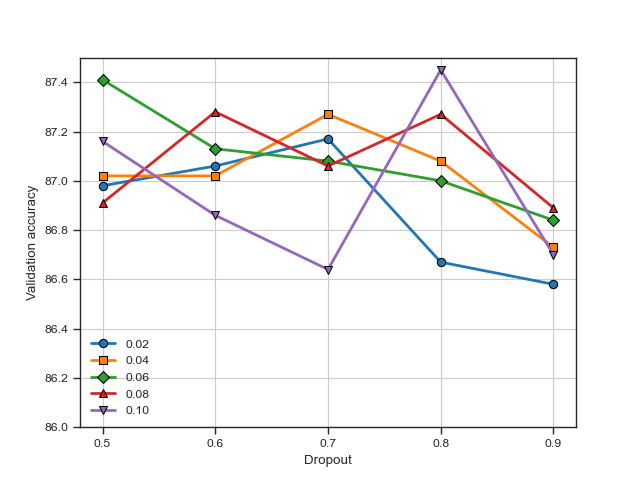}
    }
    \caption{Sentiment Classification Accuracy under Different Hyperparameter Combinations.}
    \label{fig_2}
\end{figure*}

\paragraph{First stage:} the two most important hyperparameters for DAMA are the weight of the shared auxiliary domain classification task, $\gamma$ in equation (16), and the dropout value in the neural network classification layers. Consistent with previous work, we use a grid search method, searching $\gamma$ between [0.0-0.1] with a step size of 0.02, and setting the dropout between [0.5-0.9] with a step size of 0.1. The sentiment classification accuracies on the validation set and test set under different $\gamma$ and dropout values is plotted in Fig. \ref{fig_2}.The horizontal axis corresponds to the search range of dropout, lines of different colors correspond to different $\gamma$ values, and the vertical axis represents
sentiment classification accuracy. 

According to Fig. \ref{fig_2}, we found that the optimal performance of DAMA on the validation set occurs when $\gamma$ and dropout are 0.02 and 0.5, respectively, while on the test set, the optimal choice is 0.1 and 0.8. Overall, the model performs better on the test set. These parameter choices are also those selected for BS\_mtl and DAMA, as well as for BS\_mtl (test\_set) and DAMA (test\_set). 
\begin{figure*}
    \centering
    \subfloat[On the validation set \label{fig3:sub1}]{
        \includegraphics[width=\linewidth]{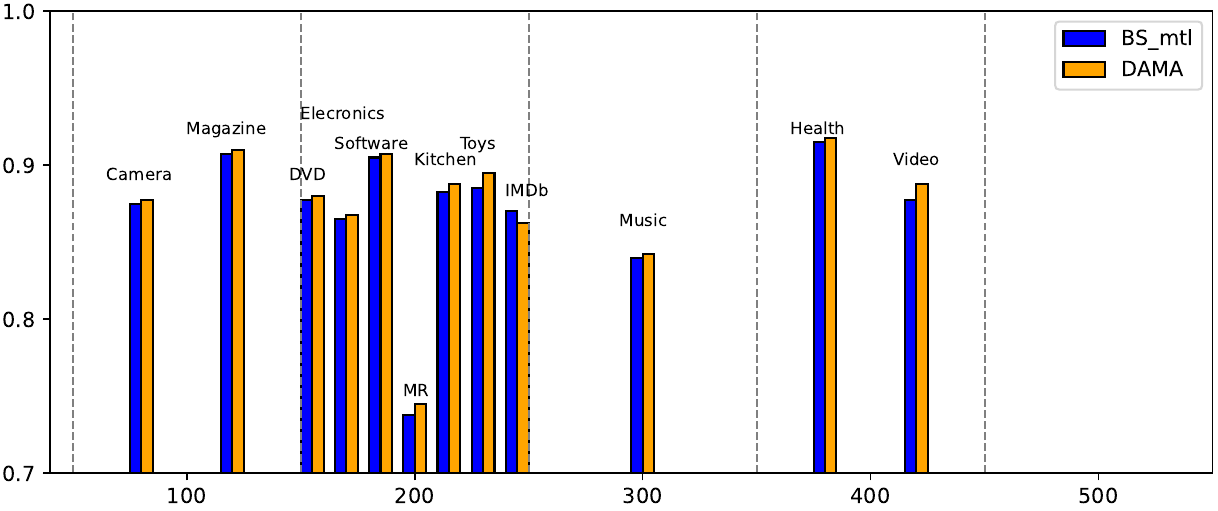}
    }
    \hfill
    \subfloat[On the test set \label{fig3:sub2}]{%
        \includegraphics[width=\linewidth]{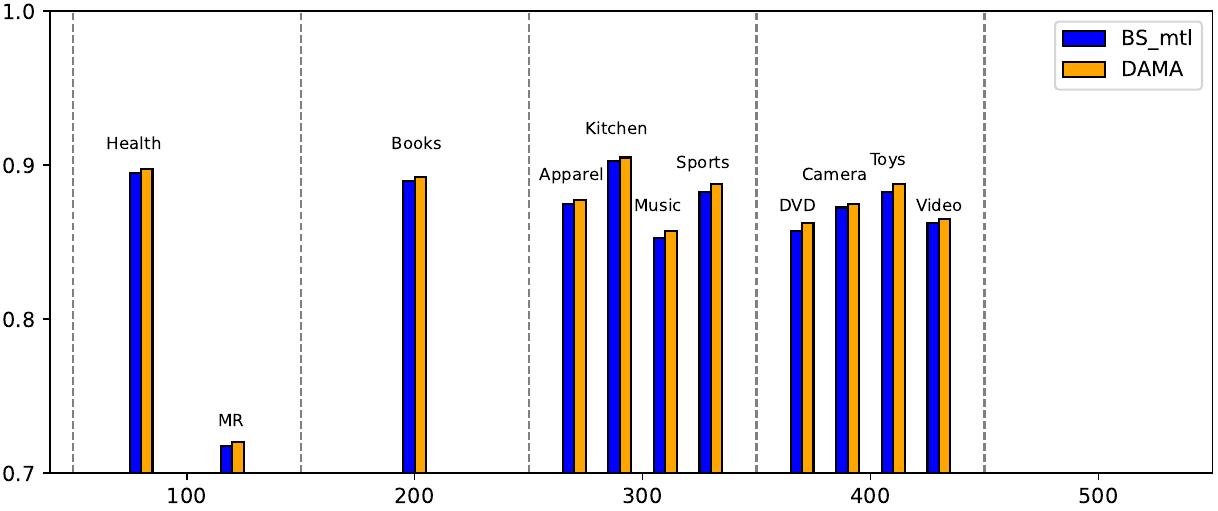}%
    }
    \caption{Selection of the range of variation for $\lambda$ for different domains}
    \label{fig_3}
\end{figure*}

\paragraph{Second stage:} the most important hyperparameter for DAMA in the second stage is the range of variation for $\lambda$. The impact of the variation should be limited between [$\gamma L^d$, $\gamma L^d$] to avoid negative domain classification loss values. In practice, we adopt a more straightforward approach\- by setting the search range between [100-500] with a step size of 100. This approach is chosen for two reasons: it avoids additional computational costs and allows us to uncover some useful conclusions. Fig. \ref{fig_3} shows the appropriate range of $\lambda$ variations across different domains on the validation and test sets. The horizontal axis corresponds to the range of variation for $\lambda$, where 100 indicates that the value of $\lambda$ is constrained within the range [-100, 100]. The vertical axis represents sentiment classification accuracy. We do not display data from domains where the classification performance is unchanged after introducing the modulation. Additionally, when the same classification accuracy is achieved across different $\lambda$ ranges, we choose the smallest value as the $\lambda$ range. 

From Fig. \ref{fig3:sub1}, it is clear that on the validation set, in the Camera and Magazine domains, DAMA obtains the most improvement when $\lambda$ is set within the range [-100, 100]. In 7 domains, such as DVD, Software and Electronics, DAMA performs best when $\lambda$ is within the range [-200, 200]. Some other domains, like Music, Health and Video, perform better when $\lambda$ is within a broader search space. Similarly, Fig. \ref{fig3:sub2} shows how the model performance can be further improved across different $\lambda$ search ranges in every domain on the test set. Comparing Fig. \ref{fig3:sub1} and Fig. \ref{fig3:sub2}, it can be observed that DAMA performs better on the validation set in most domains at $\lambda$ equals 100 or 200. However, on the test set, DAMA achieves higher performance in most domains when $\lambda$ is set to 300 or 400. This is because the shared weight for the domain classification tasks in the general models were set to 0.02 and 0.1 for the validation and test sets, respectively.. In principle, smaller initial weight values should correspond to a smaller search range of $\lambda$. Therefore, when the weight is 0.2, the optimal range of $\lambda$ for multiple domains should be narrower than when the weight is 0.1, and vice versa.

\subsection{\texorpdfstring{Changes of $\lambda$} and Classification Accuracies in DAMA}

Table \ref{tab:table2} lists the learned and adjusted values of $\lambda$ for each domain during the second stage of DAMA training, the sentiment classification accuracies of the general model on the validation set $O_{vc}$ and the test set $O_{tc}$, the sentiment classification accuracies of the model with modulation updates on these two datasets $S_{vc}$ and $S_{tc}$, as well as the domain classification accuracies $D_{vc}$ and $D_{tc}$.
\begin{table*}[!t]\centering
 \caption{Performance}
 \label{tab:table2}
 \begin{minipage}{\textwidth}\centering
 \scalebox{1}{
 \renewcommand\arraystretch{1.2}
 \begin{tabular}{|p{1.1cm}<{\centering}|p{1.9cm}<{\centering}|p{0.95cm}<{\centering}|p{1.9cm}<{\centering}|p{1.4cm}<{\centering}|p{1.9cm}<{\centering}|p{1.2cm}<{\centering}|p{1.9cm}<{\centering}|}
 \hline
\multirow{4}{*}{Books}  & $\lambda:131.6 \rightarrow 296.2$ &  \multirow{4}{*}{DVD}  & $\lambda:396.1$ &  \multirow{4}{*}{Electronics} & $\lambda:-27.1 \rightarrow-3.4$  &  \multirow{4}{*}{Apparel} & $\lambda:183.0$ \\
&$O_{vc}$: 83.1 && $O_{vc}$: 91.6 & &$O_{vc}$: 86.3 & &$O_{vc}$: 88.1 \\
&$O_{tc}$: 89.0 && $O_{tc}$: 85.8 & &$O_{tc}$: 87.3 & &$O_{tc}$: 87.5 \\
&$S_{vc}$: 83.1$\rightarrow$83.7 && $S_{vc}$: 91.6 & &$S_{vc}$: 86.0$\rightarrow$86.3 & &
$S_{vc}$: 88.4 \\
&$S_{tc}$: 89.3$\rightarrow$89.3 && $S_{tc}$: 86.3 & &$S_{tc}$: 87.3$\rightarrow$87.3 & &
$S_{tc}$: 87.8 \\
&$D_{vc}$: 0.0$\rightarrow$0.0 && $D_{vc}$: 0.3$\rightarrow$0.3 & &$D_{vc}$: 0.0$\rightarrow$0.0 & &
$D_{vc}$:  0.0$\rightarrow$0.0\\
&$D_{tc}$: 0.0$\rightarrow$0.0 && $D_{tc}$: 2.0$\rightarrow$0.8 & &$D_{tc}$: 0.0$\rightarrow$0.0 & &
$D_{tc}$: 0.0$\rightarrow$0.0\\
\hline
  \multirow{4}{*}{Software}  & $\lambda:95.8$ &  \multirow{4}{*}{Camera}  & $\lambda:286.9 \rightarrow 143.5$  &  \multirow{4}{*}{Health} & $\lambda:-31.8 \rightarrow -71.5$  &  \multirow{4}{*}{Magazine} & $\lambda:50.5$ \\
&$O_{vc}$: 88.1 && $O_{vc}$: 91.6& &$O_{vc}$: 88.7 & &$O_{vc}$: 80.2 \\
&$O_{tc}$: 90.3 && $O_{tc}$: 87.3 & &$O_{tc}$: 89.5 & &$O_{tc}$: 71.8 \\
&$S_{vc}$:88.1 && $S_{vc}$:91.3$\rightarrow$91.6& &$S_{vc}$:88.7$\rightarrow$89.1 & &
$S_{vc}$:90.1 \\
&$S_{tc}$:90.3 && $S_{tc}$:87.5$\rightarrow$87.5 & &$S_{tc}$:89.5$\rightarrow$89.8 & &
$S_{tc}$:90.3 \\
&$D_{vc}$: 0.0$\rightarrow$0.0 && $D_{vc}$: 0.0$\rightarrow$0.0& &$D_{vc}$: 28.3$\rightarrow$24.6 & & $D_{vc}$:  0.0$\rightarrow$0.0\\
&$D_{tc}$:0.0$\rightarrow$0.0 && $D_{tc}$:0.0$\rightarrow$0.0& &$D_{tc}$:30.3$\rightarrow$23.5 & & $D_{tc}$: 0.0$\rightarrow$0.0\\      
 \hline
 \multirow{4}{*}{MR}  & $\lambda:37.3 \rightarrow 9.3$ &  \multirow{4}{*}{Baby}  & $\lambda:89.7$ &  \multirow{4}{*}{Kitchen} & $\lambda:194.2$ &  \multirow{4}{*}{Music} & $\lambda:197.5 \rightarrow 296.3$ \\
&$O_{vc}$: 80.2 && $O_{vc}$: 85.7 & &$O_{vc}$: 85.7 & &
$O_{vc}$: 82.9 \\
&$O_{tc}$: 71.8 && $O_{tc}$: 88.8 & &$O_{tc}$: 90.3 & &
$O_{tc}$: 85.3 \\
&$S_{vc}$: 79.3$\rightarrow$80.2 && $S_{vc}$: 85.7& &$S_{vc}$:85.7 & &
$S_{vc}$: 82.9$\rightarrow$82.6 \\
&$S_{tc}$: 72.3$\rightarrow$72.0 && $S_{tc}$: 88.8 & &$S_{tc}$:90.5 & &
$S_{tc}$: 85.8$\rightarrow$85.8 \\
&$D_{vc}$: 100.0$\rightarrow$98.8 && $D_{vc}$: 0.0$\rightarrow$0.0& &$D_{vc}$: 0.0$\rightarrow$0.0 & &
$D_{vc}$: 9.6$\rightarrow$4.5\\
&$D_{tc}$: 100.0$\rightarrow$100.0 && $D_{tc}$: 0.0$\rightarrow$0.0& &$D_{tc}$: 0.0$\rightarrow$0.0 & &
$D_{tc}$: 3.8$\rightarrow$5.3\\   \hline
\multirow{4}{*}{Sports}  & $\lambda:-200.4$ &  \multirow{4}{*}{Toys}  & $\lambda:299.2$  &  \multirow{4}{*}{IMDb} & $\lambda:-34.2 \rightarrow -51.3$  & \multirow{4}{*}{Video} & $\lambda:360.0 \rightarrow 180.0$ \\
&$O_{vc}$: 88.2 && $O_{vc}$: 85.6 & &$O_{vc}$: 87.5 & &
$O_{vc}$: 88.9 \\
&$O_{tc}$: 88.3 && $O_{tc}$: 88.3 & &$O_{tc}$: 86.3 & &
$O_{tc}$: 86.3 \\
&$S_{vc}$: 88.7&  
&$S_{vc}$: 85.9& &$S_{vc}$: 87.5$\rightarrow$87.8 & &
$S_{vc}$: 88.6$\rightarrow$88.9 \\
&$S_{tc}$: 88.8 && $S_{tc}$:88.8 & &$S_{tc}$: 86.5$\rightarrow$86.3 & &
$S_{tc}$: 86.5$\rightarrow$86.5 \\
&$D_{vc}$: 0.0$\rightarrow$0.0 && $D_{vc}$: 0.0$\rightarrow$0.0 & &$D_{vc}$: 73.6$\rightarrow$86.5 & &
$D_{vc}$: 4.3$\rightarrow$2.0\\
&$D_{tc}$: 0.0$\rightarrow$0.0 && $D_{tc}$: 0.0$\rightarrow$0.0& &$D_{tc}$: 73.0$\rightarrow$83.8 & &
$D_{tc}$: 5.5$\rightarrow$3.3\\    
 
\hline
 \end{tabular}
}
\end{minipage}
\vspace{0mm}
\end{table*}
 
Table \ref{tab:table2} shows that in the Apparel, DVD, Software, Magazine, Baby, Kitchen, and Toys domains, $\lambda$ did not require adjustment through the scaling strategy. This indicates that the $\lambda$  learned via the Adam algorithm brings improvements on the validation set compared to the general model, except in the DVD domain, where scaling would have caused $\lambda$ to exceed the search range, so it remained unchanged. For example, in the Apparel domain, the classification accuracy of the general model $O_{vc}$ is 88.1\%, while in DAMA, the classification accuracy $S_{vc}$ is 88.4\%. In the Books, Health, Music, and IMDb domains, $\left|\lambda\right|$ increased after scaling. This suggests that DAMA did not improve on the validation set with the learned $\lambda$, hence a larger step size was used. For instance, in the Books domain, the classification accuracies of both the general model and DAMA $O_{vc}$ are 83.1\%. In other domains, $\left|\lambda\right|$ decreased. This indicates that the performance of DAMA on the validation set degraded with the learned $\lambda$, so it should be reduced by a constant factor. For example, in the Electronics domain, the classification accuracy of the general model $O_{vc}$ is 86.3\%, while in DAMA, the classification accuracy $S_{vc}$ is 86.0\%.

Next, by comparing the trends of $S_{vc}$ and $S_{tc}$, it is found that in most cases, after $S_{vc}$ improves, $S_{tc}$ either remains the same or improved. Similarly, the trends of $D_{vc}$ and $D_{tc}$ are generally consistent. This demonstrated that our proposed model has good generalization ability.

Additionally, it can be seen that through joint learning, sentiment classification tasks are heavily dependent on the domain classification tasks in some specific domains. For example, in the MR domain, the domain classification accuracy reaches 100\%. In IMDb, Health, DVD, Music and Video domains, relatively high domain classification accuracies are attained. In other domains, this value is 0.0\%, although the auxiliary weight of the domain classification task is set to 0.02 in the general model. It proves that the essence of the auxiliary domain classification task is not for domain classification, but to provide the domain information needed for the sentiment classification task. Therefore, even though in some domains, such as Toys, the domain classification accuracy remains at 0.0\%, the performance of sentiment classification is enhanced. Another point worth noting is that when the initial domain classification accuracy is not zero, changes in $\lambda$ will directly affect the model's domain classification performance. However, its‘ sign and change cannot determine whether the domain classification accuracy will increase or decrease, as this also needs to be decided in conjunction with the gradient.

\subsection{Extended  Experiments}
In the previous discussion, we mentioned that loss-based  and gradient-based multi-task learning algorithms face certain challenges. This section displays the performance of three naïve  models, \-BS\_scale, BS\_PCGrad1 and BS\_PCGrad2 on the mtl\-data in Table \ref{tab:table3}. 

\begin{table}[ht]
\caption{Experimental results on the other baselines\label{tab:table3}}
    \centering
    \renewcommand\arraystretch{1.2}
    \begin{tabular}{|c|c|} \hline 
    \textbf{Model} & \textbf{Sentiment classification Performance}  \\ \hline
    BS\_scale & Test Accuracy: 50.3~$\rightarrow$50.5 \\
             &~~~~~~~~~~~~~~~~~~~~~~~~$\rightarrow$50.5 \\
             &~~~~~~~~~~~~~~~~~~~~~~~~$\rightarrow$50.5 \\
             &~~~~~~~~~~~~~~~~~~~~~~~~$\rightarrow$49.5  \\ \hline
    BS\_PCGrad1 &	Batch size: $\geq$2, 0ut of Memory  	 \\ \hline
    BS\_PCGrad2 & Batch size: $\geq$2, 0ut of Memory  \\ \hline
    \end{tabular}
\end{table}
As observed from Table \ref{tab:table3}, in multi-domain sentiment classification scenarios, jointly training $\gamma^j$ results in the model failing to converge. The classification accuracy of BS\_scale remains around 50\% through 5 epochs training. Moreover, in our experimental environment, BS\_PCGrad1 and BS\_PCGrad2 were unable to run because gradient-based methods require significant memory to store gradients, which poses serious challenges in terms of computational resources.

\subsection{Further Discusssion}
Our work adopts a multi-task learning framework as the basic model in which all tasks are considered equally important - that is, the aim of training in each domain is to achieve the highest sentiment classification accuracy as soon as possible. This task characteristic allows us to further decouple the original joint training problem into a set of domain-specific performance optimization problems. Such a perspective offers new insights for many similar frameworks, such as multimodal learning. These methods typically rely on single-stage training, where feature extraction often involves a game-theoretic process to achieve optimal or suboptimal trade-off solutions. However, we argue that beyond this, one should also consider whether the extracted features for each task require additional filtering or supplementation. Moreover, in the context of multi-domain sentiment analysis, as the number of domains or tasks increases, joint training approaches face significant challenges due to the rapidly expanding search space, which often leads to only suboptimal or even inferior solutions. In such cases, our proposed second-stage optimization becomes increasingly important.

\section{Conclusion}
In this work, we propose a novel dynamic domain information modulation algorithm to improve the performance of multi-domain sentiment analysis models. To solve the problem of dynamic balance between domain classification tasks, we introduce domain-aware and sample-aware modulations based on a multi-task learning architecture, where the original one-stage joint training approach is expanded into a two-stage process. In the first stage, a shared weight for domain classification tasks is obtained, while in the second stage, modulation is introduced to refine domain-specific sentiment classification by adjusting domain information based on single-domain data, which helps to produce more interpretable and stable experimental results. 

Furthermore, we propose a method to calculate the modulations based on loss-based and gradient-based methods. Every modulation is divided into two components: step size and direction, so that it changes along the gradient of the domain classification loss, and the classic Adam algorithm is used to learn the step size to minimize the sentiment classification loss in each domain. This method based on input signal update, especially only updating a scalar step size for each domain instead of adjusting all network parameters, can effectively reduce the computing resources and time cost for model training. To avoid overfitting over steps, we further design a step scaling strategy based on the validation set. The reduction and maintenance of classification performance on the validation set are associated with excessive and negligible modulations, and the step size in each domain is adjusted accordingly. Comprehensive experimental results on multi-domain sentiment analysis datasets with 16 domains indicate that our proposed models outperform baseline models in overall performance when considering classification accuracy, time and algorithmic complexity.

Meanwhile, we also recognize that there are still many avenues worth exploring for the proposed method. First, our method introduces new hyperparameters. These hyperparameters use conventional grid search and some group-based heuristic search methods may help the model learn the step size more efficiently. Second, we can further analyze the relationship between the gradients and step sizes contained in the modulations in each domain, reveal the relationships between the sentiment classification tasks on each domain, and then build a model to handle gradient conflicts problems, maximizing mutual benefits. These are very promising research topics.


\bibliographystyle{unsrtnat}
\bibliography{references}

\end{document}